\documentclass{article}

\PassOptionsToPackage{numbers,square,sort&compress}{natbib}
\usepackage{graphicx}
\usepackage{multirow}
\usepackage{wrapfig}
\usepackage{colortbl}
\usepackage{amsmath, amssymb}
\usepackage{booktabs}
\usepackage{wrapfig}
\usepackage{multirow}
\usepackage{subcaption}
\usepackage{caption}

\usepackage[preprint]{corl_2026} % Uncomment for pre-prints (e.g., arxiv); This is like ``final'', but will remove the CORL footnote.

\title{DexSynRefine: Synthesizing and Refining Human--Object Interaction Motions for Physically Feasible Dexterous Robot Actions}

\author{
  Hyesung Lee$^{1,2,*}$,
  Hyunwoo Jung$^{3}$,
  Si-Hwan Heo$^{1}$, 
  Sungwook Yang$^{1}$ \\
  \\
  $^{1}$Korea Institute of Science and Technology, $^{2}$KAIST, $^{3}$Hanyang University \\
  \url{https://dexsynrefine.github.io/}
}

\begin{document}
\maketitle

%===============================================================================
% ABSTRACT
% \begin{abstract}
% Learning dexterous manipulation from human-object interaction (HOI) data offers a scalable alternative to robot teleoperation, but HOI demonstrations are sparse and purely kinematic, making direct retargeting unreliable under embodiment mismatch and contact-rich dynamics. We present \textbf{DexSynRefine}, a coupled framework that treats HOI data as structured motion priors rather than executable commands. DexSynRefine synthesizes task- and initial-object-state-conditioned hand-object trajectories with HOI-MMFP, physically grounds them through task-space residual reinforcement learning, and infers missing contact-and-dynamics context from proprioceptive history for deployment. Across five dexterous manipulation tasks, each stage addresses a complementary bottleneck: HOI-MMFP improves trajectory consistency and smoothness, task-space residuals provide the most effective grounding representation, and contact-dynamics adaptation enables real-world execution. Together, these stages improve real-world success rates over kinematic retargeting by $50$--$70$ percentage points.
% \end{abstract}

\begin{abstract}
Learning dexterous manipulation from human--object interaction (HOI) data offers a scalable alternative to robot teleoperation, but HOI demonstrations are typically sparse and purely kinematic, making direct retargeting unreliable under embodiment mismatch and contact-rich dynamics. We present \textbf{DexSynRefine}, a coupled framework that treats HOI data as structured motion priors rather than executable robot actions. DexSynRefine first synthesizes hand--object trajectories conditioned on the task and initial object state using HOI Motion Manifold Flow Primitives (HOI-MMFP), a motion prior for coupled hand--object motion. It then physically grounds them with task-space residual reinforcement learning and adapts execution by inferring missing contact--dynamics context from proprioceptive history. Across five dexterous manipulation tasks, each stage addresses a complementary bottleneck: HOI-MMFP improves trajectory consistency and smoothness, task-space residuals provide the strongest grounding representation among the tested alternatives, and contact--dynamics adaptation enables robust real-world execution. Together, DexSynRefine improves real-world success rates over kinematic retargeting by $50$--$70$~percentage points.
\end{abstract}


%===============================================================================
%% 1. INTRODUCTION
\section{Introduction}
\label{SECTION:SECTION1}
% figure 1 %%
% \begin{wrapfigure}{r}{0.5\textwidth}
%     \vspace{-0.3in}
%     \centering
%     \includegraphics[width=0.48\textwidth]{Figures/figure_1.pdf}
%     \caption{\textbf{DexSynRefine} synthesizes HOI references from sparse demonstrations and physically grounds them into executable dexterous robot actions.}
%     \label{FIG:FIGURE_1}
%     \vspace{-20pt}
% \end{wrapfigure}

%% Figure 1 %%

\begin{wrapfigure}[18]{r}[0pt]{0.44\textwidth}
    \vspace{-0.20in}
    \centering
    \includegraphics[width=\linewidth]{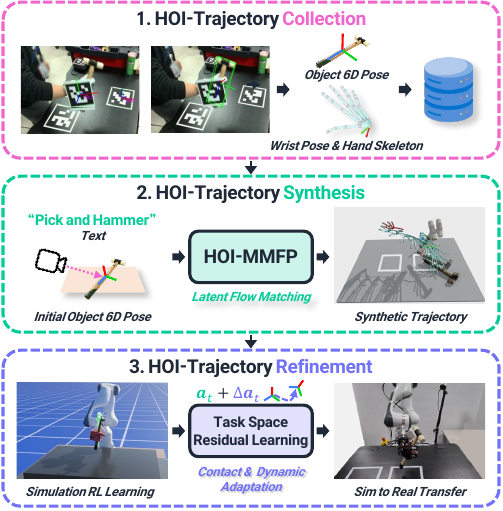}
    \vspace{-6pt}
    \caption{\textbf{DexSynRefine} synthesizes HOI references from sparse demonstrations and grounds them into robot actions.}
    \label{FIG:FIGURE_1}
    \vspace{-4pt}
\end{wrapfigure}

Recent advances in imitation learning~\cite{chi2023diffusion, mandlekar2021matters, zhao2023learning, heng2025vitacformer} have significantly improved multi-fingered dexterous manipulation, enabling increasingly dexterous and contact-rich robot behaviors learned from large-scale teleoperation datasets~\cite{khazatsky2024droid, o2024open, zhao2025humanoid, cheng2024open}. However, collecting such datasets remains expensive, labor-intensive, and difficult to scale, especially for dexterous tasks that require coordinated finger motion, accurate contact timing, and stable object control. Because human hands naturally produce rich contact patterns during everyday object interaction, a growing line of work has turned to more accessible recordings of human-object interaction (HOI)---ranging from Internet videos to motion-capture sessions~\cite{chao2021dexycb, fan2023arctic, banerjee2024introducing, lu2025humoto}---as a scalable source of motion priors.

Yet HOI demonstrations are not robot demonstrations, and converting them into executable dexterous robot behavior remains difficult for two reasons. First, HOI data are inherently sparse: a few recorded interactions cannot cover the range of task configurations or initial object states encountered at deployment, so a suitable human reference may not exist for a new object pose. Second, HOI data are purely kinematic: it provides geometric hand and object motion, while successful manipulation depends on unobserved contact forces, friction, actuation limits, and other contact-rich dynamics. This difficulty is further amplified by the embodiment gap between human and robot hands, which differ in kinematic structure, actuation, and contact geometry. Recent work has mitigated part of this gap by using human or object motion as references for goal-conditioned reinforcement learning~\cite{chen2024object, li2025maniptrans, zhao2024dexh2r, chen2024vividex, liu2025dextrack, lum2025crossing}. However, these methods still assume an appropriate reference interaction for each task instance. The key challenge is therefore to synthesize task- and initial-object-state-conditioned interaction priors from sparse HOI data, physically ground these priors into feasible robot actions, and make this grounding deployable without privileged simulation states.

We address these coupled challenges with \textbf{DexSynRefine}, a framework that treats sparse HOI data as structured motion priors and converts them into deployable dexterous robot actions. First, DexSynRefine synthesizes task- and initial-object-state-conditioned interaction trajectories using HOI-MMFP, an extension of Motion Manifold Flow Primitives (MMFP)~\cite{lee2024mmp++, lee2025motion} to coupled hand-object motion, constructing references beyond sparse demonstrations while preserving the observed interaction structure. Second, the synthesized trajectories are physically grounded through task-space residual reinforcement learning, which refines wrist and fingertip targets into feasible robot behavior under embodiment mismatch and contact-rich dynamics. Finally, real-robot deployment introduces an additional challenge: object properties and contact states available in simulation are not directly observable from onboard sensing. While prior adaptation methods such as RMA~\cite{qi2023hand, liang2024rapid} infer hidden dynamics from proprioceptive history, dexterous manipulation also requires rapid responses to finger-object contact transitions; DexSynRefine therefore couples contact estimation with dynamics adaptation from proprioceptive history, enabling deployable residual refinement without privileged simulation states.

Our contributions are threefold: (1) HOI-MMFP, a task- and initial-object-state-conditioned generative prior for coherent hand-object trajectory synthesis from object-centric sparse-demonstration augmentation; (2) a task-space residual RL formulation for physically grounding synthesized HOI references under embodiment and contact-dynamics constraints, validated through a systematic action-representation study; and (3) a contact-and-dynamics adaptation module that infers missing physical context from proprioceptive history for deployable real-world dexterous manipulation.

% Our contributions are threefold: (1) HOI-MMFP, a task- and initial-object-state-conditioned generative prior that synthesizes coherent hand--object trajectories from sparsely collected and object-centrically augmented HOI demonstrations; (2) a task-space residual RL formulation for physically grounding synthesized HOI references under embodiment and contact-dynamics constraints, validated through a systematic action-representation study; and (3) a contact-and-dynamics adaptation module that infers missing physical context from proprioceptive history for deployable real-world dexterous manipulation.

%===============================================================================
%% 2. RELATED WORKS
\section{Related Work}
\label{SECTION:SECTION2}

\paragraph{HOI Priors and Reference-Guided Reinforcement Learning.}
A promising direction for bridging kinematic human motion and dynamic robot execution is to use human-object interaction (HOI) demonstrations as motion priors and learn absolute or residual policies through reinforcement learning. Existing methods mainly differ in their action representations: Zhao et al.~\cite{zhao2024dexh2r} learn joint-space residuals around retargeted primitives; Li et al.~\cite{li2025maniptrans} fine-tune a residual policy after imitation pretraining; Chen et al.~\cite{chen2024object} combine task-space wrist residuals with absolute finger commands in a hierarchical policy; and Lum et al.~\cite{lum2025crossing} learn absolute joint actions on top of a geometric fabric controller, using human hand kinematics only during pre-manipulation.

These works show that human motion provides useful structure for reinforcement learning, but leave two issues unresolved. First, they generally assume a suitable reference interaction for each deployment instance, which becomes restrictive under sparse HOI data or unseen initial object poses. Second, despite using different absolute or residual action spaces, the role of action representation in HOI-guided physical grounding remains underexplored. DexSynRefine addresses these gaps by synthesizing task- and initial-object-state-conditioned HOI references with HOI-MMFP and systematically comparing grounding representations. Our results identify task-space residuals over wrist and fingertip targets as an effective formulation for preserving interaction structure while correcting embodiment- and contact-induced infeasibility.

%===============================================================================
%% 3. METHOD
\section{Method}
\label{SECTION:METHOD}

\subsection{Task Setup}
\label{SUBSECTION:TASK_SETUP}
We consider the problem of converting sparse human-object interaction (HOI) demonstrations into executable robot behavior for dexterous manipulation. Each demonstration is represented as
\begin{equation}
\boldsymbol{\tau} = \{(\mathbf{w}_t, \mathbf{h}_t, \mathbf{o}_t)\}_{t=1}^{T},
\end{equation}
where $\mathbf{w}_t \in \mathrm{SE}(3)$ and $\mathbf{o}_t \in \mathrm{SE}(3)$ denote the wrist pose and object pose in the world frame, respectively, and $\mathbf{h}_t \in \mathbb{R}^{K \times 3}$ denotes the $K$ hand keypoints expressed in the wrist frame. Each trajectory is associated with a conditioning variable $\mathbf{c} = (\mathbf{u}, \mathbf{o}_0)$, where $\mathbf{u}$ is a task embedding and $\mathbf{o}_0$ is the initial object pose.

Given a sparse HOI demonstration set $\mathcal{D} = \{(\boldsymbol{\tau}^{(i)}, \mathbf{c}^{(i)})\}_{i=1}^{N}$, our goal is to synthesize task-consistent HOI references for unseen initial object poses and physically ground them into executable robot actions. DexSynRefine addresses this problem in three stages: (1) object-centric augmentation of sparse HOI trajectories, (2) task- and initial-object-state-conditioned HOI motion prior synthesis via HOI-MMFP, and (3) residual policy learning with deployable contact-and-dynamics adaptation.

\subsection{HOI Motion Collection and Object-Centric Augmentation}
Because each task is represented by only a few HOI demonstrations, the collected trajectories cannot cover the range of initial object poses required at deployment. We collect seven HOI demonstrations per task for five dexterous manipulation tasks (\textit{Reorient Pringles Bottle, Pick Up and Hammer, Pick and Pour Watering Can, Pick and Place Book, Pick and Place Bowl}) spanning pick-and-place, tool use, and object reorientation, using a multi-camera setup with FoundationPose~\cite{wen2024foundationpose} and Manus Gloves, while varying the object's initial yaw across demonstrations.

To expand pose coverage while preserving the interaction structure, we apply object-centric augmentation: initial object poses are perturbed by planar $\mathrm{SE}(2)$ transformations, with the reaching phase interpolated and the manipulation phase rigidly transformed to preserve relative hand-object geometry. For symmetric objects, we additionally randomize rotations around the symmetry axis. This yields approximately $300$ augmented trajectories per task, each resampled to a fixed horizon of $T=220$. Details are provided in Appendix~\ref{appendix:APPENDIX_A}.

\subsection{HOI-MMFP: Conditional HOI Motion Prior Synthesis}
\begin{figure*}[t]\centering
\vspace{-8pt}
    \centering
    \includegraphics[width = \textwidth]{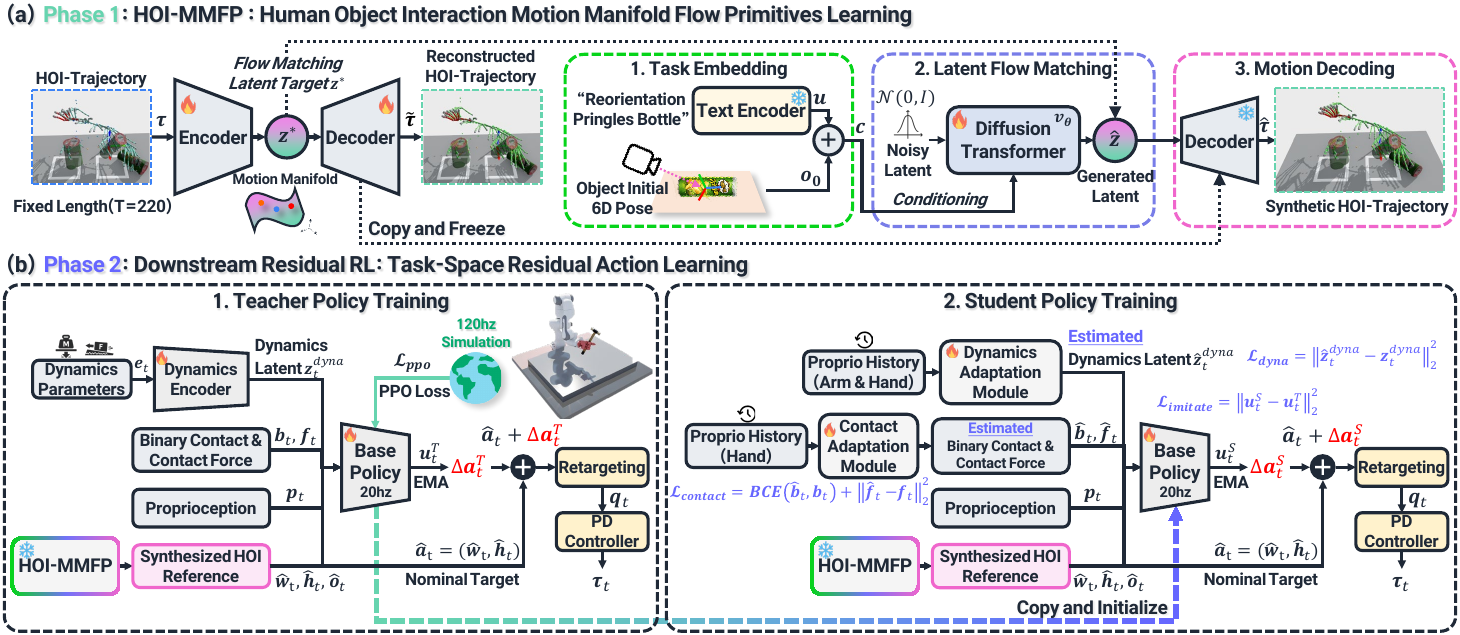} %max. 7.16 inch
    \caption{Overview of \textbf{DexSynRefine}. 
    \textbf{(a) HOI-MMFP}: an autoencoder learns a compact HOI motion manifold, and a conditional flow-matching model generates task- and initial-object-state-conditioned latents decoded into synthesized HOI references. 
    \textbf{(b) Task-Space Residual RL}: a privileged teacher policy physically grounds the synthesized reference in simulation, and a deployable student policy infers contact and dynamics information from proprioceptive histories.}\label{FIG:FIGURE_2}
    \vspace{-8pt}
\end{figure*}

Full-trajectory generation is high-dimensional and prone to temporal jitter, while simple conditional latent models may lose spatial grounding under unseen object poses. HOI-MMFP addresses this by extending Motion Manifold Flow Primitives (MMFP)~\cite{lee2024mmp++, lee2025motion} to coupled hand-object motion: it learns a compact HOI motion manifold and performs conditional flow matching in latent space. The training pipeline is shown in Figure~\ref{FIG:FIGURE_2}(a).

HOI-MMFP is trained in two stages: manifold learning and conditional latent flow matching. Given an augmented trajectory $\boldsymbol{\tau}^{\mathrm{aug}}$, a Transformer autoencoder maps it to a latent code $\mathbf{z}^{*} = E(\boldsymbol{\tau}^{\mathrm{aug}})$ and reconstructs it as $\tilde{\boldsymbol{\tau}} = D(\mathbf{z}^{*})$. The autoencoder is optimized with
\begin{equation}
    \mathcal{L}_{\mathrm{AE}} = \mathcal{L}_{\mathrm{recon}} + \eta \mathcal{L}_{\mathrm{norm}} + \delta \mathcal{L}_{\mathrm{smooth}},
\end{equation}
where $\mathcal{L}_{\mathrm{recon}} = \|\boldsymbol{\tau}^{\mathrm{aug}} - \tilde{\boldsymbol{\tau}}\|_2^2$ and $\mathcal{L}_{\mathrm{norm}} = \|\mathbf{z}^{*}\|_2^2$. To encourage manifold continuity, $\mathcal{L}_{\mathrm{smooth}}$ penalizes temporal differences of decoded interpolated latents $\boldsymbol{\tau}_{\mathrm{mix}} = D(\alpha \mathbf{z}^{*} + (1-\alpha) \mathbf{z}_j)$, where $\mathbf{z}_j$ is a randomly permuted latent code and $\alpha \sim \mathcal{U}(\alpha_{\min}, \alpha_{\max})$.

For task- and initial-object-state-conditioned generation, we train a latent flow-matching model using the conditioning variable $\mathbf{c} = (\mathbf{u}, \mathbf{o}_0)$ defined in Sec.~\ref{SUBSECTION:TASK_SETUP}, where $\mathbf{u}$ is obtained from a pretrained text encoder. Let $\mathbf{z}_0 \sim \mathcal{N}(\mathbf{0}, \mathbf{I})$ and $\mathbf{z}_s = (1-s)\mathbf{z}_0 + s\mathbf{z}^{*}$ for $s \sim \mathcal{U}(0,1)$. A Diffusion Transformer~\cite{peebles2023scalable}-based vector field $v_{\theta}$ is trained with
\begin{equation}
\mathcal{L}_{\mathrm{FM}} =
\mathbb{E}_{\boldsymbol{\tau}^{\mathrm{aug}}, \mathbf{z}_0, s}
\left[
\left\|
v_{\theta}(\mathbf{z}_s, s, \mathbf{c}) - (\mathbf{z}^{*} - \mathbf{z}_0)
\right\|_2^2
\right].
\end{equation}
At inference, we solve the ODE $\frac{\mathrm{d}\mathbf{z}_s}{\mathrm{d}s} = v_{\theta}(\mathbf{z}_s, s, \mathbf{c})$ from $s=0$ to $s=1$, decode $\hat{\mathbf{z}}$ as $\hat{\boldsymbol{\tau}} = D(\hat{\mathbf{z}})$, and obtain $\hat{\boldsymbol{\tau}} = \{(\hat{\mathbf{w}}_t, \hat{\mathbf{h}}_t, \hat{\mathbf{o}}_t)\}_{t=1}^{T}$ as a task-consistent kinematic prior for downstream physical grounding.

%% figure 2 %%

\subsection{Task-Space Residual Refinement with Contact-and-Dynamics Adaptation}
The synthesized HOI reference is kinematically plausible but not directly executable under embodiment mismatch and contact-rich constraints. We therefore learn a task-space residual policy that locally corrects the synthesized wrist and fingertip targets while preserving the hand-object interaction structure, as shown in Figure~\ref{FIG:FIGURE_2}(b). At each episode start, we randomize the initial object pose, query HOI-MMFP once to obtain $\hat{\boldsymbol{\tau}}$, and reuse this fixed reference throughout the rollout.

\paragraph{Task-Space Residual Action Integration.}
At each policy step $t$, the refinement policy predicts a task-space residual signal $\mathbf{u}_t$, smoothed by an exponential moving average (EMA):
\begin{equation}
\tilde{\mathbf{u}}_t = \alpha \mathbf{u}_t + (1 - \alpha) \tilde{\mathbf{u}}_{t-1},
\end{equation}
where $\alpha$ is the smoothing coefficient. The residual action $\Delta \mathbf{a}_t$ (denoted $\Delta \mathbf{a}^T_t$ for the teacher and $\Delta \mathbf{a}^S_t$ for the student) is obtained by integrating the filtered command and clipping it to predefined bounds:
\begin{equation}
\Delta \mathbf{a}_t = \text{clip}\!\left( \Delta \mathbf{a}_{t-1} + s \cdot \tilde{\mathbf{u}}_t \cdot \Delta t, \; \Delta \mathbf{a}_{\text{min}}, \; \Delta \mathbf{a}_{\text{max}} \right),
\end{equation}
with scaling factor $s$ and $\Delta t = 0.05$\,s. Clipping keeps the correction bounded around the synthesized HOI reference. The refined task-space target is obtained by adding $\Delta \mathbf{a}_t$ to the wrist and fingertip components of $\hat{\boldsymbol{\tau}}_t$.

To bridge the $20$\,Hz policy and $120$\,Hz simulation, we apply quintic task-space interpolation and solve inverse kinematics---damped least-squares for the 7-DoF arm and an analytical solver for the 16-DoF hand---to obtain joint commands $\mathbf{q}_t$ at $120$\,Hz.

\paragraph{Privileged Teacher Policy Learning.}
To learn physically grounded corrections, we first train a privileged teacher policy in simulation using PPO~\cite{schulman2017proximal} with an asymmetric actor-critic formulation. The critic observes the privileged tuple $ (\mathbf{s}_t^{\mathrm{priv}}, \mathbf{p}_t^{\mathrm{hand}}, \mathbf{p}_t^{\mathrm{arm}}, \mathbf{b}_t, \mathbf{f}_t, \hat{\boldsymbol{\tau}}_t, \mathbf{e}_t) $, including simulator-only states, proprioception, contact states and forces, the synthesized reference, and object dynamics parameters. Detailed dimensions are provided in Appendix~\ref{appendix:APPENDIX_C}.

The actor receives a dynamics latent $\mathbf{z}_t^{\mathrm{dyna}} = \mu(\mathbf{e}_t)$ and predicts $\mathbf{u}_t^T$ from $(\mathbf{z}_t^{\mathrm{dyna}}, \mathbf{p}_t^{\mathrm{hand}}, \mathbf{p}_t^{\mathrm{arm}}, \mathbf{b}_t, \mathbf{f}_t, \hat{\boldsymbol{\tau}}_t)$. This asymmetric design lets the critic exploit richer simulator information while the actor refines the synthesized reference using structured kinematic, contact, and dynamics cues. The reward includes object tracking, fingertip and wrist guidance, contact encouragement, and penalties on unintended collisions and excessive residual changes.

\paragraph{Deployable Student Contact-and-Dynamics Adaptation.}
Since privileged contact and dynamics variables are unavailable on the real robot, we distill a student policy that infers this missing physical context from proprioceptive history. We maintain a hand-centric history $\mathbf{H}_t^{\mathrm{hand}}$ over $\{\mathbf{p}_{t-k}^{\mathrm{hand}}\}_{k=0}^{K-1}$ and a full-body history $\mathbf{H}_t^{\mathrm{full}}$ over $\{[\mathbf{p}_{t-k}^{\mathrm{arm}}; \mathbf{p}_{t-k}^{\mathrm{hand}}]\}_{k=0}^{K-1}$ with $K=30$.

A GRU~\cite{chung2014empirical}-based contact module predicts $(\hat{\mathbf{b}}_t, 
\hat{\mathbf{f}}_t) = \phi_{\mathrm{con}}(\mathbf{H}_t^{\mathrm{hand}})$, while a dynamics module infers 
$\hat{\mathbf{z}}_t^{\mathrm{dyna}} = \phi_{\mathrm{dyn}}(\mathbf{H}_t^{\mathrm{full}})$. 
The student actor, initialized from the teacher, predicts $\mathbf{u}_t^S$ from $(\hat{\mathbf{z}}_t^{\mathrm{dyna}}, \mathbf{p}_t^{\mathrm{hand}}, \mathbf{p}_t^{\mathrm{arm}}, \hat{\mathbf{b}}_t, \hat{\mathbf{f}}_t, \hat{\boldsymbol{\tau}}_t)$ and is trained with
\begin{equation}
\mathcal{L}_{\mathrm{student}} =
\mathcal{L}_{\mathrm{imitate}} +
\mathcal{L}_{\mathrm{dyna}} +
\mathcal{L}_{\mathrm{contact}},
\end{equation}
where $\mathcal{L}_{\mathrm{imitate}} = \|\mathbf{u}_t^S - \mathbf{u}_t^T\|_2^2$, 
$\mathcal{L}_{\mathrm{dyna}} = \|\hat{\mathbf{z}}_t^{\mathrm{dyna}} - 
\mathbf{z}_t^{\mathrm{dyna}}\|_2^2$, and $\mathcal{L}_{\mathrm{contact}} 
= \lambda_b \mathrm{BCE}(\hat{\mathbf{b}}_t, \mathbf{b}_t) + \lambda_f 
\|\hat{\mathbf{f}}_t - \mathbf{f}_t\|_2^2$. This enables deployable residual refinement from proprioceptive history alone.

%===============================================================================
%% 4. EXPERIMENTS AND RESULTS
\section{Experiments and Results}
\label{SECTION:EXPERIMENTS}
We design our experiments to validate the key bottlenecks in converting sparse HOI data into executable dexterous robot actions: reference synthesis, physical grounding, deployable adaptation, and real-world transfer. Specifically, we ask four questions: \textbf{(Q1)} Can HOI-MMFP synthesize spatially consistent and smooth HOI references that are useful for downstream physical refinement? \textbf{(Q2)} Which action representation best grounds synthesized HOI references into physically feasible robot behavior? \textbf{(Q3)} Are contact and dynamics adaptation jointly necessary for deployable residual refinement? \textbf{(Q4)} Does the full DexSynRefine pipeline enable real-world dexterous robot execution from synthesized HOI references?

We evaluate DexSynRefine across five dexterous manipulation tasks. Unless otherwise specified, simulation results are averaged over 100 environments, 10 trials, and 10 random seeds. Performance metrics include success rate (SR) at termination, object translation error (OTE, in meters), and object orientation error (OOE, in degrees), where OTE and OOE are averaged over the entire episode.

\subsection{Motion Synthesis Quality and Downstream RL Usability}
\label{sec:exp_motion_synthesis}

\paragraph{Baselines.}
We compare HOI-MMFP against a Task-Conditioned VAE (TC-VAE)~\cite{noseworthy2020tcvae}, which generates trajectories by combining a task-independent latent representation with explicit task conditioning, and a full-trajectory variant (DiT-Full) that applies our flow-matching backbone directly in trajectory space, ablating the learned motion manifold. Full architectural and training details for all three priors are provided in Appendix~\ref{appendix:APPENDIX_B}. We evaluate (i) spatial consistency under initial-pose perturbations, and (ii) temporal smoothness and downstream RL utility.

\begin{figure*}[t]
    \centering
    \centering
    \includegraphics[width = \textwidth]{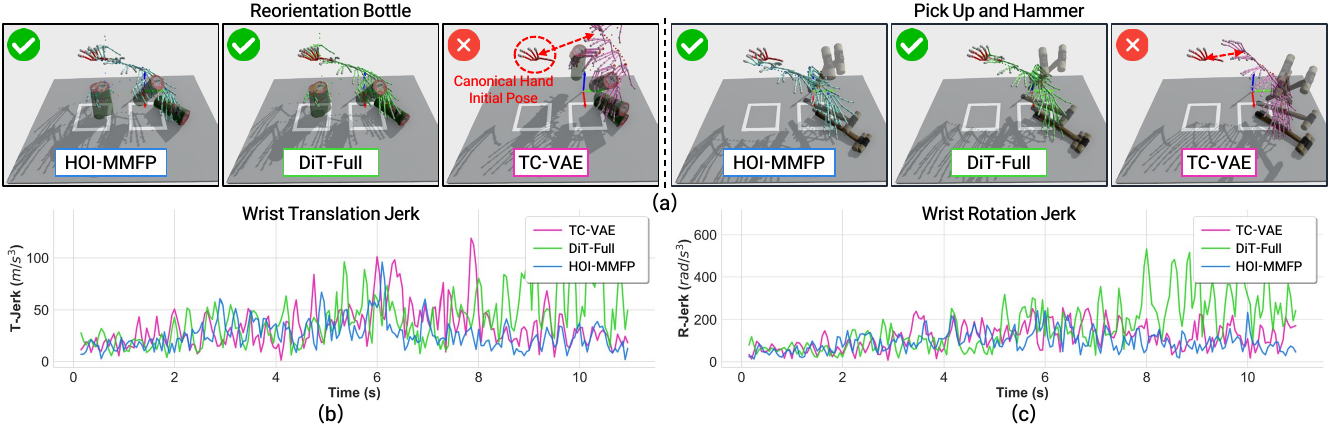} %max. 7.16 inch
    \caption{Qualitative comparison of HOI synthesis models. 
(a) Generated object and wrist trajectories visualized for two tasks. 
(b, c) Translational and rotational wrist jerk over time on the 
\textit{Pick Up and Hammer} task. Full results are provided in 
Figure~\ref{FIG:FIGURE_7}.}\label{FIG:FIGURE_3}
\vspace{-8pt}
\end{figure*}

\paragraph{Spatial Perturbation Robustness.}
Since our augmentation enforces a canonical initial hand configuration, a well-trained prior should preserve this canonical start regardless of object placement. We sweep the initial object pose along $x,y$ ($\pm 20$\,cm, $1$\,cm resolution) and yaw ($\pm 30^\circ$, $5^\circ$ resolution), and measure the first-frame wrist error against the canonical reference. As shown in Table~\ref{TABLE:TABLE_1}, HOI-MMFP yields the lowest error ($0.015$\,m / $2.50^\circ$), outperforming DiT-Full ($0.018$\,m / $2.65^\circ$) and substantially improving over TC-VAE ($0.121$\,m / $12.88^\circ$). Figure~\ref{FIG:FIGURE_3}(a) further shows that HOI-MMFP and DiT-Full preserve a consistent canonical start under out-of-distribution placements, while TC-VAE produces distorted trajectories with shifted starting poses, indicating a loss of spatial grounding.

%%--------------------------------------------------------------
%% figure 3 %%

%%--------------------------------------------------------------
%% table 1 %%
\begin{table}[t]
\setlength{\abovecaptionskip}{4pt}
\setlength{\belowcaptionskip}{4pt}
\centering
\scriptsize
\begin{tabular*}{\linewidth}{@{\extracolsep{\fill}} l cc ccc @{}}
\toprule
 & \multicolumn{2}{c}{Trajectory Prior (avg.)} & \multicolumn{3}{c}{Downstream Residual RL} \\
\cmidrule(lr){2-3}\cmidrule(lr){4-6}
Method & Pos Err $\downarrow$ & Rot Err $\downarrow$ & SR (\%) $\uparrow$ & OTE $\downarrow$ & OOE $\downarrow$ \\
\midrule
TC-VAE ~\cite{noseworthy2020tcvae}            & $0.121{\scriptstyle\pm 0.068}$ & $12.88{\scriptstyle\pm 11.02}$ & --- & --- & --- \\
DiT-Full        & $0.018{\scriptstyle\pm 0.010}$ & $2.65{\scriptstyle\pm 1.54}$   & $49.10{\scriptstyle\pm 4.12}$ & $0.061{\scriptstyle\pm 0.010}$ & $55.42{\scriptstyle\pm 2.54}$ \\
HOI-MMFP (Ours) & $\mathbf{0.015{\scriptstyle\pm 0.010}}$ & $\mathbf{2.50{\scriptstyle\pm 1.71}}$ & $\mathbf{60.30{\scriptstyle\pm 3.04}}$ & $\mathbf{0.053{\scriptstyle\pm 0.003}}$ & $\mathbf{50.00{\scriptstyle\pm 1.39}}$ \\
\bottomrule
\end{tabular*}
\caption{Synthesized HOI reference quality (left, averaged over 5 tasks) and downstream residual RL performance on \textit{Pick Up and Hammer} (right).}
\label{TABLE:TABLE_1}
\vspace{-8pt}
\end{table}

\paragraph{Trajectory Smoothness and Downstream Utility.}
We assess temporal smoothness using the mean-squared jerk of the wrist pose, computed over the trajectories in Figure~\ref{FIG:FIGURE_3}(a) and shown in Figure~\ref{FIG:FIGURE_3}(b), (c). HOI-MMFP yields the lowest translational and rotational jerk ($29.83{\scriptstyle\pm 4.78}$\,m/s$^3$, $101.08{\scriptstyle\pm 17.12}$\,rad/s$^3$), outperforming TC-VAE ($37.00{\scriptstyle\pm 8.52}$\,/\,$139.14{\scriptstyle\pm 45.21}$) and DiT-Full ($41.03{\scriptstyle\pm 8.23}$\,/\,$189.52{\scriptstyle\pm 33.82}$). DiT-Full is particularly prone to high-jerk motion when generating directly in full trajectory space without a manifold constraint. In simulation on \textit{Pick Up and Hammer}, a representative contact-rich task, HOI-MMFP also outperforms DiT-Full in downstream residual RL (SR $60.3\%$ vs.\ $49.1\%$, OTE $0.053$ vs.\ $0.061$\,m, OOE $50.0^\circ$ vs.\ $55.4^\circ$; Table~\ref{TABLE:TABLE_1} right). TC-VAE is excluded from downstream RL because its severe geometric distortions prevent convergence to a non-trivial success rate. Thus, manifold-based synthesis provides HOI references that are both spatially grounded and smooth enough for downstream physical refinement.

\subsection{Comparing Action Representations for Physical Grounding}
\label{sec:exp_action_repr}

\paragraph{Baselines.}
Holding the synthesized HOI reference and reward design fixed, we compare five action representations: (i) \textbf{Joint-Space Absolute}, (ii) \textbf{Joint-Space Residual}, (iii) \textbf{Task-Space Absolute}, (iv) \textbf{Task-Space Residual (Ours)}, and (v) \textbf{Object-Centric Dexterous}, a hybrid baseline following~\cite{chen2024object} that applies task-space residuals to the wrist while predicting absolute joint commands for the fingers. We also include a \textbf{Kinematic Retargeting} baseline that executes the synthesized HOI reference through inverse kinematics without learned refinement, isolating the need for physical grounding. Detailed formulations of each representation are provided in Appendix~\ref{appendix:APPENDIX_D}.

\begin{figure*}[t]
    \centering
    \centering
    \includegraphics[width = \textwidth]{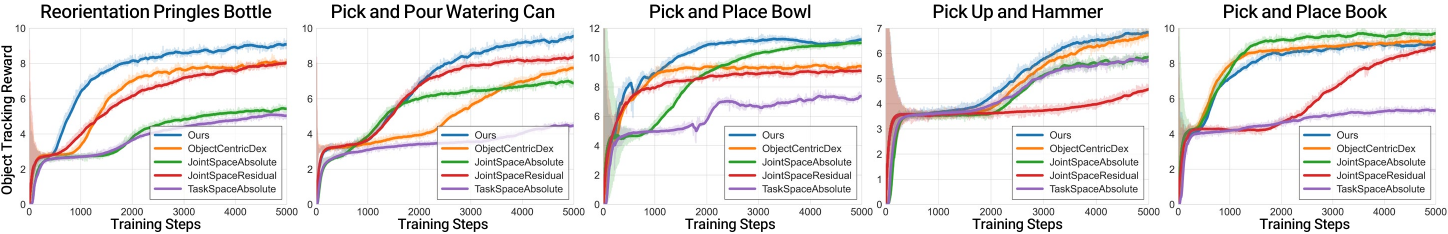} %max. 7.16 inch
    \caption{Object tracking reward over training steps across five tasks.}\label{FIG:FIGURE_4}
\end{figure*}

\paragraph{Results.}
Table~\ref{TABLE:TABLE_2} reports SR, OTE, and OOE, while Figure~\ref{FIG:FIGURE_4} shows object-tracking reward curves. Kinematic retargeting nearly fails across all tasks (SR $\leq 5.8\%$), confirming that synthesized HOI references are not directly executable. Absolute baselines also struggle: \textit{Task Abs} fails universally by destabilizing wrist-object alignment, while \textit{Joint Abs} succeeds mainly on simple pick-and-place tasks (\textit{Book}, \textit{Bowl}). Although \textit{Object-Centric} is competitive, our \textbf{Task-Space Residual} achieves the highest mean SR ($68.1\%$) and performs best on all contact-rich tasks (\textit{Pringles}, \textit{Hammer}, \textit{Watering Can}). While \textit{Joint Abs} slightly outperforms ours on \textit{Book} (SR $64.4\%$ vs.\ $61.6\%$, OTE $0.035$ vs.\ $0.054$, OOE $20.0^\circ$ vs.\ $24.7^\circ$) due to its larger direct action space in low-contact regimes, it collapses on contact-rich tasks ($\leq 13\%$ SR). These results indicate that physical grounding is most effective when the policy locally corrects task-space wrist and fingertip targets instead of regenerating the full motion.

%%--------------------------------------------------------------
%% figure 4 %%
%%--------------------------------------------------------------

%%--------------------------------------------------------------
%% table 2 %%
\begin{table}[t]
\vspace{-8pt}  
\setlength{\abovecaptionskip}{4pt}
\setlength{\belowcaptionskip}{4pt}
\centering
\scriptsize
\setlength{\tabcolsep}{2.5pt}
\begin{tabular}{l ccc ccc ccc ccc ccc}
\toprule
& \multicolumn{3}{c}{Pringles} & \multicolumn{3}{c}{Watering Can} & \multicolumn{3}{c}{Bowl} & \multicolumn{3}{c}{Hammer} & \multicolumn{3}{c}{Book} \\
\cmidrule(lr){2-4}\cmidrule(lr){5-7}\cmidrule(lr){8-10}\cmidrule(lr){11-13}\cmidrule(lr){14-16}
Action & SR & OTE & OOE & SR & OTE & OOE & SR & OTE & OOE & SR & OTE & OOE & SR & OTE & OOE \\
\midrule
Kinematic Retar.        & $5.8$  & $0.120$ & $88.8$  & $0.0$  & $0.300$ & $65.1$ & $2.2$  & $0.147$ & $20.0$ & $0.0$  & $0.190$ & $101.8$ & $0.0$  & $0.283$ & $64.8$ \\
Joint Abs              & $1.4$  & $0.119$ & $39.1$  & $4.1$  & $0.130$ & $40.9$ & $89.0$ & $0.032$ & $10.8$ & $13.0$ & $0.151$ & $87.9$  & $\mathbf{64.4}$ & $\mathbf{0.035}$ & $\mathbf{20.0}$ \\
Joint Res              & $51.4$ & $0.071$ & $43.4$  & $43.9$ & $0.113$ & $\mathbf{27.5}$ & $7.1$  & $0.092$ & $24.3$ & $1.0$  & $0.169$ & $84.7$  & $41.5$ & $0.081$ & $30.5$ \\

Task Abs               & $0.0$  & $0.134$ & $78.5$  & $0.0$  & $0.132$ & $51.0$ & $23.1$ & $0.143$ & $28.4$ & $18.0$ & $0.131$ & $65.1$  & $0.5$  & $0.144$ & $41.9$ \\
Object-Centric~\cite{chen2024object} & $66.4$ & $0.050$ & $33.3$ & $19.5$ & $0.155$ & $50.3$ & $57.7$ & $0.055$ & $21.1$ & $52.7$ & $0.083$ & $55.4$ & $60.4$ & $0.045$ & $23.8$ \\
\midrule
\textbf{Task Res (Ours)} & $\mathbf{71.5}$ & $\mathbf{0.044}$ & $\mathbf{26.5}$ & $\mathbf{52.4}$ & $\mathbf{0.107}$ & $34.2$ & $\mathbf{94.8}$ & $\mathbf{0.023}$ & $\mathbf{6.8}$ & $\mathbf{60.3}$ & $\mathbf{0.053}$ & $\mathbf{50.0}$ & $61.6$ & $0.054$ & $24.7$ \\
\bottomrule
\end{tabular}
\caption{Comparison of action representations on five tasks.}
\label{TABLE:TABLE_2}
\vspace{-8pt}
\end{table}
%%--------------------------------------------------------------

%%--------------------------------------------------------------
\subsection{Effectiveness of Contact-and-Dynamics Adaptation}
\label{sec:exp_adaptation}

\paragraph{Baselines.}
To evaluate whether contact and dynamics provide complementary physical context, we perform ablations on the two most contact-rich tasks---\textit{Pick Up and Hammer} and \textit{Pick and Pour Watering Can}. As shown in Table~\ref{TABLE:TABLE_3}, we compare our \textbf{Full (Ours)} adaptation against two variants: \textbf{w/o Contact}, which adapts only dynamics and is equivalent to standard RMA~\cite{qi2023hand}, and \textbf{w/o Dynamics}, which adapts only contact.

\paragraph{Results.}
The \textbf{Full} teacher achieves the highest success rate and lowest errors while both ablations nearly fail, showing that contact and dynamics are jointly necessary for residual refinement. Dynamics adaptation captures object-level physical variations, whereas contact adaptation captures rapid finger-object interaction changes. During distillation, the \textbf{Full} student remains the best overall, nearly matching the teacher on \textit{Watering Can} ($51.8\%$ vs.\ $52.4\%$), but showing a larger gap on \textit{Hammer} ($44.3\%$ vs.\ $60.3\%$), where pre-swing realignment involves more complex contact transitions that are harder to infer from proprioceptive history alone.

%%--------------------------------------------------------------
%% figure 5 %%
\begin{figure*}[t]
    \centering
    \centering
    \includegraphics[width = \textwidth]{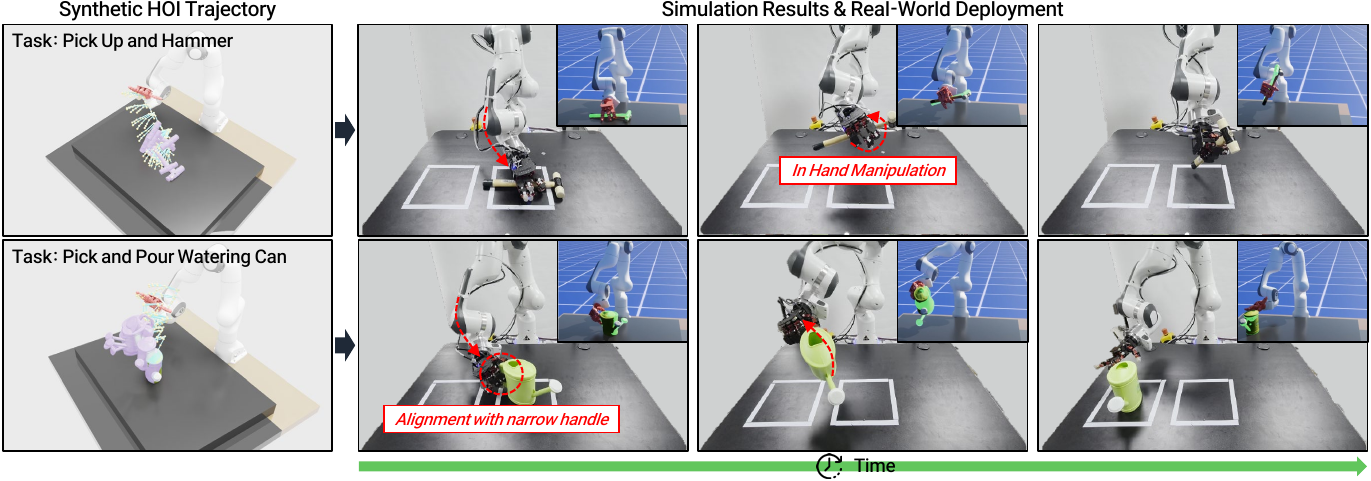} %max. 7.16 inch
    \caption{\textbf{Real-world and simulation deployment.} Synthesized HOI references (left) and corresponding simulation and real-robot executions (right), shown for \textit{Pick Up and Hammer} (top) and \textit{Pick and Pour Watering Can} (bottom). See Figure~\ref{FIG:FIGURE_10} for remaining tasks.}\label{FIG:FIGURE_5}
\vspace{-8pt}
\end{figure*}
%%--------------------------------------------------------------

%%--------------------------------------------------------------
%% table 3 & 4 %%
\begin{table}[t]
\setlength{\abovecaptionskip}{2pt}
\setlength{\belowcaptionskip}{2pt}
\noindent
% ===== Left: Ablation =====
\begin{minipage}[t]{0.65\linewidth}
\raggedright
\scriptsize
\begin{tabular*}{\linewidth}{@{\extracolsep{\fill}} l l ccc ccc @{}}
\toprule
& & \multicolumn{3}{c}{Hammer} & \multicolumn{3}{c}{Watering Can} \\
\cmidrule(lr){3-5}\cmidrule(lr){6-8}
Stage & Adaptation & SR & OTE & OOE & SR & OTE & OOE \\
\midrule
\multirow{3}{*}{Teacher}
& \textbf{Full (Ours)}        & $\mathbf{60.3}$ & $\mathbf{0.053}$ & $\mathbf{50.0}$ & $\mathbf{52.4}$ & $\mathbf{0.107}$ & $\mathbf{34.2}$ \\
& w/o Contact & $23.4$          & $0.152$          & $87.2$          & $0.1$           & $0.243$ & $72.2$          \\
& w/o Dynamics & $9.3$           & $0.168$          & $97.0$          & $5.6$           & $0.219$ & $67.0$          \\
\midrule
\multirow{3}{*}{Student}
& \textbf{Full (Ours)}        & $\mathbf{44.3}$ & $\mathbf{0.068}$ & $\mathbf{54.7}$          & $\mathbf{51.8}$ & $\mathbf{0.111}$ & $\mathbf{36.4}$ \\
& w/o Contact & $17.2$          & $0.163$          & $92.5$ & $0.0$           & $0.228$          & $73.4$          \\
& w/o Dynamics & $7.5$           & $0.175$          & $103.6$          & $4.4$           & $0.223$          & $67.8$          \\
\bottomrule
\end{tabular*}
\captionof{table}{Adaptation ablation study on the \textit{Pick Up and Hammer} and \textit{Pick and Pour Watering Can} tasks.}
\label{TABLE:TABLE_3}
\end{minipage}\hspace{6pt}%
% ===== Right: Real-world =====
\begin{minipage}[t]{0.33\linewidth}
\raggedright
\scriptsize
\renewcommand{\arraystretch}{1.55}
\begin{tabular*}{\linewidth}{@{\extracolsep{\fill}} l cc @{}}
\toprule
Task & Kin.\ Retarget. & \textbf{DexSynRefine} \\
\midrule
Pringles  & $2/10$ & $\mathbf{8/10}$ \\
Hammer     & $1/10$ & $\mathbf{7/10}$ \\
Watering Can  & $0/10$ & $\mathbf{5/10}$ \\
Bowl         & $2/10$ & $\mathbf{9/10}$ \\
Book         & $0/10$ & $\mathbf{5/10}$ \\
\bottomrule
\end{tabular*}
\captionof{table}{Real-world success comparison.}
\label{TABLE:TABLE_4}
\end{minipage}
\vspace{-13pt}
\end{table}
%%--------------------------------------------------------------

%%--------------------------------------------------------------
\subsection{Real-World Performance Evaluation}
\label{sec:exp_real}

\paragraph{Real-World Experimental Setup.}
We evaluate the full DexSynRefine pipeline on a real robot across all five tasks, reporting success rates over $10$ trials per task. Our platform consists of a 7-DoF arm with a 16-DoF dexterous hand running closed-loop at $20$\,Hz. The initial object pose required by HOI-MMFP is estimated by FoundationPose from a calibrated RGB-D camera and queried once at episode start. We compare DexSynRefine against direct kinematic retargeting, which executes the synthesized HOI-MMFP reference through inverse kinematics without learned residual refinement or deployable adaptation.

\paragraph{Results.}
Table~\ref{TABLE:TABLE_4} reports real-world success rates across all five tasks. Kinematic retargeting achieves at most $2/10$ successful trials and fails completely ($0/10$) on \textit{Watering Can} and \textit{Book}, showing that synthesized HOI references alone are insufficient for real-world deployment. In contrast, the full DexSynRefine pipeline grounds these references through residual refinement and contact-and-dynamics adaptation, yielding absolute improvements of $50$--$70$ percentage points across tasks. DexSynRefine reaches $9/10$ success on \textit{Bowl}, $8/10$ on \textit{Pringles}, and transfers to all five tasks. The remaining failures---finger clearance failure during release (\textit{Watering Can}) and toppling during upright reorientation (\textit{Book})---indicate that contact clearance and dynamic balancing remain the main sources of sim-to-real error.
%%--------------------------------------------------------------

%===============================================================================
%% 5. Conclusion
\section{Conclusion}
\label{sec:conclusion}
We presented \textbf{DexSynRefine}, a framework for converting sparse, kinematic HOI demonstrations into executable dexterous robot behavior. DexSynRefine combines three coordinated stages: HOI-MMFP synthesizes task- and initial-object-state-conditioned HOI references, task-space residual RL physically grounds them under embodiment and contact-dynamics constraints, and contact-and-dynamics adaptation enables deployment from proprioceptive history without privileged simulation states. Across five dexterous manipulation tasks spanning pick-and-place, tool use, and object reorientation, DexSynRefine produces smoother and more spatially consistent references than synthesis baselines, identifies task-space residuals as an effective grounding representation, and improves real-world success rates over kinematic retargeting by $50$--$70$ percentage points. Together, these results show that sparse human motion priors can become executable dexterous robot actions when motion synthesis, physical grounding, and deployable adaptation are jointly considered.

%===============================================================================
%% 6. Limitations
\section{Limitations}
\label{sec:limitations}
We identify three key limitations of DexSynRefine. \textbf{(i) Per-task residual policies.} DexSynRefine currently focuses on state generalization within each task: HOI-MMFP synthesizes references for unseen initial object poses, while the residual policy is trained per task to handle task-specific contact dynamics. Extending the residual stage to a single multi-task policy is an important direction for future work. \textbf{(ii) Lossy contact reconstruction at deployment.} Lacking tactile sensors, the student infers contact states $(\hat{\mathbf{b}}_t, \hat{\mathbf{f}}_t)$ from proprioception alone. This can miss rapid contact transitions, especially in contact-rich tasks such as \textit{Hammer}, where accurate pre-swing realignment is critical. Tactile sensing could reduce this gap by directly observing fingertip contact transitions. \textbf{(iii) Episode-start-only conditioning of the prior.} HOI-MMFP synthesizes $\hat{\boldsymbol{\tau}}$ only once from the initial observation $\mathbf{o}_0$. If the object is displaced mid-execution, the synthesized reference may become inconsistent with the current object state, and the system cannot reactively re-plan. A natural extension is to periodically update the HOI prior using online perception, enabling re-planning under object perturbations.
%===============================================================================

\clearpage
% The acknowledgments are automatically included only in the final and preprint versions of the paper.
% \acknowledgments{If a paper is accepted, the final camera-ready version will (and probably should) include acknowledgments. All acknowledgments go at the end of the paper, including thanks to reviewers who gave useful comments, to colleagues who contributed to the ideas, and to funding agencies and corporate sponsors that provided financial support.}

%===============================================================================

\bibliography{reference}  % .bib

%===============================================================================
% Appendix START
%===============================================================================

\appendix
\section*{Appendix}

% =====================================================================
% Appendix A
% =====================================================================
\section{Details of HOI Data Collection and Trajectory Augmentation}
\label{appendix:APPENDIX_A}

\subsection{HOI Data Collection System}
\begin{figure*}[h]
    \centering
    \centering
    \includegraphics[width = \textwidth]{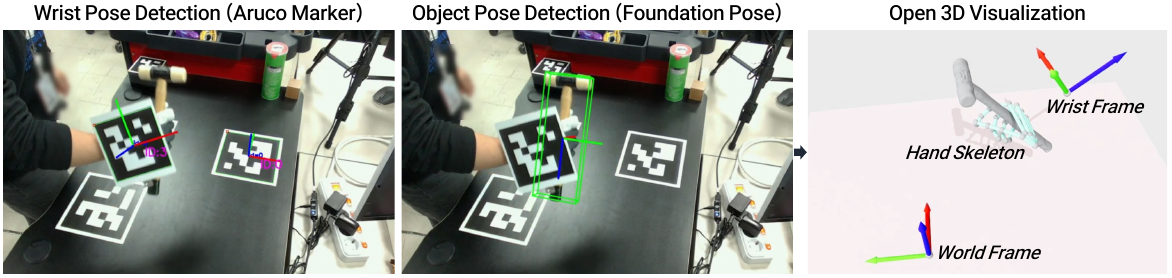} %max. 7.16 inch
    \caption{Overview of the HOI data collection system. }\label{FIG:FIGURE_6}
\vspace{-8pt}
\end{figure*}

Figure~\ref{FIG:FIGURE_6} shows our HOI data collection setup and the corresponding Open3D visualization. We attached three ArUco markers to the table to define a fixed world frame. For wrist tracking, an ArUco marker cube was attached to the wrist region of the Manus glove. Object poses were estimated from two external cameras using FoundationPose and transformed into the shared world frame. The object meshes required by FoundationPose were acquired using a smartphone-based 3D scanner app, which provides high-fidelity textured meshes from a brief handheld scan of each object.

The calibration aims to estimate the fixed rigid transformation from the Manus hand skeleton frame $S$ to the wrist marker cube frame $W$, denoted as ${}^{W}T_{S}$. For this, we collected approximately 30 static samples with different hand poses. For each sample $j$, we logged the fingertip position $p^{S}_{j}$ in the Manus skeleton frame, the corresponding calibration point $p^{C}_{j}$ in the camera frame $C$, and the wrist cube pose ${}^{C}T_{W}(j)$. The camera-observed point was first transformed into the wrist cube frame:
\begin{equation}
p^{W}_{j} =
\left({}^{C}T_{W}(j)\right)^{-1} p^{C}_{j}.
\end{equation}

We then solve a robust rigid registration problem:
\begin{equation}
\operatorname*{arg\,min}_{R,t}
\sum_{j=1}^{N}
\rho \left(
\left\|
p^{W}_{j} - \left(R p^{S}_{j} + t\right)
\right\|^{2}
\right).
\end{equation}
where $R$ and $t$ define ${}^{W}T_{S}$, and $\rho(\cdot)$ is a robust loss. The optimization is initialized using SVD-based rigid alignment. Outliers are removed based on the initial residuals, and the transformation is refined using IRLS with a Huber loss. The final calibration accuracy is validated using RMSE between the transformed Manus fingertip positions and the corresponding points in the wrist cube frame.

%% figure 6 %%

\subsection{Object-Centric Trajectory Augmentation}
We augment sparse HOI demonstrations in an object-centric manner to improve generalization to unseen initial object poses. 
We follow the notation in Sec.~\ref{SUBSECTION:TASK_SETUP}, where $\mathbf{w}_t$, $\mathbf{o}_t$, and $\mathbf{h}_t$ denote the wrist pose, object pose, and hand keypoints in the wrist frame, respectively. 
The corresponding world-frame hand keypoints are computed as
\begin{equation}
\mathbf{x}_{t,k}
=
R(\mathbf{w}_t)\mathbf{h}_{t,k}
+
p(\mathbf{w}_t),
\qquad
k=1,\dots,K.
\end{equation}

We first remove the redundant pre-contact segment. 
The first interaction frame is detected by the wrist-object distance:
\begin{equation}
t_c
=
\min_t
\left\{
\|p(\mathbf{o}_t)-p(\mathbf{w}_t)\| < \delta_c
\right\}.
\end{equation}
All frames before $t_c$ are discarded.

To generate diverse initial object poses, we sample a planar object-centric transformation
\begin{equation}
\mathbf{G}
=
\begin{bmatrix}
R_z(\theta) & \mathbf{b} \\
0 & 1
\end{bmatrix},
\end{equation}
where $R_z(\theta)$ is a yaw rotation about the world $z$-axis. 
Given the initial object position $\mathbf{p}_0=p(\mathbf{o}_1)$ and a sampled planar translation $\mathbf{d}=[d_x,d_y,0]^\top$, we set
\begin{equation}
\mathbf{b}
=
\mathbf{p}_0 + \mathbf{d} - R_z(\theta)\mathbf{p}_0 .
\end{equation}
The wrist and object trajectories are transformed together:
\begin{equation}
\mathbf{w}'_t = \mathbf{G}\mathbf{w}_t,
\qquad
\mathbf{o}'_t = \mathbf{G}\mathbf{o}_t .
\end{equation}
Since the same transformation is applied to both poses, the relative hand-object geometry is preserved.

The reaching phase is regenerated from a fixed canonical hand start pose $\mathbf{w}_{\mathrm{start}}$ to the first augmented interaction pose $\mathbf{w}'_1$:
\begin{equation}
\mathbf{w}^{\mathrm{reach}}_{\alpha}
=
\mathrm{Interp}
\left(
\mathbf{w}_{\mathrm{start}},
\mathbf{w}'_1;
\alpha
\right),
\qquad
\alpha \in [0,1].
\end{equation}
Translation is linearly interpolated, while rotation is interpolated using slerp. 
The hand keypoints in the wrist frame are also linearly interpolated from the canonical initial hand configuration to $\mathbf{h}_1$.

For the object-use phase, we preserve the original object-to-wrist relation:
\begin{equation}
\mathbf{T}^{ow}_t
=
\mathbf{o}_t^{-1}\mathbf{w}_t .
\end{equation}
Given the augmented object trajectory $\hat{\mathbf{o}}_t$, the wrist trajectory is reconstructed as
\begin{equation}
\hat{\mathbf{w}}_t
=
\hat{\mathbf{o}}_t
\mathbf{T}^{ow}_t .
\end{equation}
This retargets the hand motion to the augmented object pose while maintaining the original grasp relation.

For axis-symmetric objects, we additionally apply a random local-axis rotation to the object orientation:
\begin{equation}
R(\hat{\mathbf{o}}_t)
\leftarrow
R(\hat{\mathbf{o}}_t)R_a(\phi),
\qquad
\phi \sim \mathcal{U}(-\pi,\pi),
\end{equation}
where $R_a(\phi)$ is a rotation around the object's local symmetry axis. 
Finally, all augmented trajectories are resampled to a fixed length using linear interpolation for translations and hand keypoints, and slerp for rotations.

% =====================================================================
% Appendix B
% =====================================================================
\section{Details of Generative Motion Prior Baselines}
\label{appendix:APPENDIX_B}

This section describes the implementation of HOI-MMFP and the two generative motion-prior baselines (TC-VAE and DiT-Full) used in Sec.~\ref{sec:exp_motion_synthesis}. To isolate the effect of \emph{where} generation happens (latent vs.\ trajectory space) and \emph{how} the prior is trained (one-stage vs.\ two-stage), all three methods share the same trajectory representation, conditioning interface, and training schedule; method-specific differences are restricted to the generative formulation.

% --------------------------------------------------------------------
\subsection{Shared Setup}
\label{appendix:B-APPENDIX_B.1}

\paragraph{Trajectory representation.}
Each episode is a fixed-length sequence
$\boldsymbol{\tau}\!\in\!\mathbb{R}^{T\times D}$ with $T{=}220$ and per-frame
features $D{=}93$:
$[\,\mathbf{w}^{\text{pos}}_t\,(3),\ \mathbf{w}^{\text{rot6d}}_t\,(6),\
\mathbf{o}^{\text{pos}}_t\,(3),\ \mathbf{o}^{\text{rot6d}}_t\,(6),\
\mathbf{h}^{\text{local}}_t\,(75)\,]$.
Rotations use the 6D representation; trajectories are normalized
per-feature using training-set statistics.

\paragraph{Conditioning.}
All methods receive the same conditioning variable
$\mathbf{c}=(\mathbf{u},\mathbf{o}_0)$. The numeric component
$\mathbf{o}_0\!\in\!\mathbb{R}^{9}$ is the initial object pose
(3D position + 6D rotation), and the semantic component
$\mathbf{u}\!\in\!\mathbb{R}^{384}$ is obtained from the frozen
\texttt{all-MiniLM-L6-v2} encoder~\cite{reimers2019sentence}
(mean-pooled, $\ell_2$-normalized). Inside each network,
$\mathbf{o}_0$ and $\mathbf{u}$ are projected by separate two-layer MLPs
and added to the conditioning vector.

\paragraph{Training.}
We use identical optimization for all methods: fused AdamW
(lr $3{\times}10^{-4}$, wd $10^{-4}$, grad clip $1.0$), 1{,}000 epochs,
batch size $256$, bf16 mixed precision, and the same random seed and
batch ordering. The augmented HOI dataset (canonical wrist start,
randomized object $x,y,$ yaw) is split 95/5 into train/val. The shared
backbone uses hidden dim $d_h{=}256$ with $3$ Transformer layers and
$4$ attention heads.

% --------------------------------------------------------------------
\subsection{HOI-MMFP}
\label{appendix:B-APPENDIX_B.2}

\paragraph{Stage 1: Motion manifold AE.}
A 3-layer Transformer encoder ($d_h{=}256$, 4 heads, GELU, dropout 0.1) maps $\boldsymbol{\tau}^{\mathrm{aug}}$ to $\mathbf{z}^{*}\!\in\!\mathbb{R}^{16}$ via mean pooling and a LayerNorm--Linear head. The decoder broadcasts $\mathbf{z}^{*}$ over $T{=}220$ time tokens (after a 2-layer MLP expansion), adds positional embeddings, and applies a 3-layer Transformer followed by a linear head. Loss weights: $\eta{=}10^{-3}$, $\delta{=}10^{-2}$, $\alpha\!\sim\!\mathcal{U}(-0.4,1.4)$.

\paragraph{Stage 2: Latent conditional flow.}
With the AE frozen, a velocity field $v_\theta$ operates on a sequence of
$16$ scalar tokens (one per latent coordinate), each embedded to
$d_h{=}256$. Three adaLN-Zero DiT blocks~\cite{peebles2023scalable}
(4 heads, MLP ratio 4) are modulated by the sum of (i) a sinusoidal time
embedding (dim $256$, scale $10^{3}$), (ii) projected $\mathbf{o}_0$, and
(iii) projected $\mathbf{u}$. Inference uses a 50-step Heun ODE solver to
integrate $\mathrm{d}\mathbf{z}_s/\mathrm{d}s
{=}v_\theta(\mathbf{z}_s,s,\mathbf{c})$ from $s{=}0$ to $s{=}1$, after
which $\hat{\boldsymbol{\tau}}{=}D(\hat{\mathbf{z}})$.

% --------------------------------------------------------------------
\subsection{TC-VAE Baseline}
\label{appendix:B-APPENDIX_B.3}

We follow the original Task-Conditioned VAE formulation~\cite{noseworthy2020tcvae}: the latent $\mathbf{h}\!\in\! \mathbb{R}^{16}$ encodes \emph{manner} variation that must be statistically independent of the specified task condition $\mathbf{c}$. We replace the original TCN backbone with a Transformer encoder--decoder that matches HOI-MMFP in capacity, and keep the rest of the recipe unchanged. The encoder $q_\phi(\mathbf{h}|\boldsymbol{\tau})$ outputs a diagonal Gaussian, and the decoder $p_\theta(\boldsymbol{\tau}|\mathbf{h},\mathbf{c})$ is conditioned on both $\mathbf{h}$ and $\mathbf{c}$. Training combines a $\beta$-VAE ELBO with an adversarial information objective: an auxiliary 3-layer MLP $f_W$ predicts $\mathbf{c}$ from $\boldsymbol{\mu}_\phi(\boldsymbol{\tau})$, and the encoder is trained to fool it,
\begin{equation}
\min_{\theta,\phi}\,\mathcal{L}_{\mathrm{ELBO}}
- \alpha\,\mathcal{L}_{\mathrm{aux}},
\qquad
\min_{W}\,\mathcal{L}_{\mathrm{aux}}
=\big\|f_W(\boldsymbol{\mu}_\phi(\boldsymbol{\tau}))-\mathbf{c}\big\|_{1},
\end{equation}
where
$\mathcal{L}_{\mathrm{ELBO}}
=\beta\,\mathrm{KL}(q_\phi\|\mathcal{N}(0,I)) -\mathbb{E}_q[\log p_\theta]$. Following~\cite{noseworthy2020tcvae} we set $\beta{=}8$ and tune $\alpha$ on the validation set. At inference we draw $\mathbf{h}\!\sim\!\mathcal{N}(0,I)$ and decode under $\mathbf{c}$. Because reconstruction does not enforce any condition-equivariant structure on the latent, the decoder must re-derive spatial grounding from $\mathbf{c}$ alone, which is brittle under out-of-distribution $\mathbf{o}_0$ (cf.\ Table~\ref{TABLE:TABLE_1}).

% --------------------------------------------------------------------
\subsection{DiT-Full Baseline}
\label{appendix:B-APPENDIX_B.4}

DiT-Full ablates the manifold by applying our flow-matching backbone directly to the raw trajectory: tokens are time steps rather than latent coordinates, and the input $\mathbf{x}_s\!\in\!\mathbb{R}^{T\times D}$ is linearly projected to $d_h{=}256$ with $T{=}220$ learned positional embeddings. Three DiT blocks (4 heads, MLP ratio 4) modulated by the same $(s,\mathbf{o}_0,\mathbf{u})$ produce a per-frame velocity $v_\theta\!\in\!\mathbb{R}^{T\times D}$. Training uses the rectified flow-matching loss with $\mathbf{x}_s{=}(1{-}s)\mathbf{x}_0+s\boldsymbol{\tau}^{\mathrm{aug}}$, and inference uses the same 50-step Heun integrator. DiT-Full and HOI-MMFP therefore differ \emph{only} in the space the flow operates on, providing a controlled ablation of the learned motion manifold.

% --------------------------------------------------------------------
\subsection{Qualitative Results and Latent Structure}
\label{appendix:B-APPENDIX_B.5}

\begin{figure*}[h]
    \centering
    \includegraphics[width=0.7\textwidth]{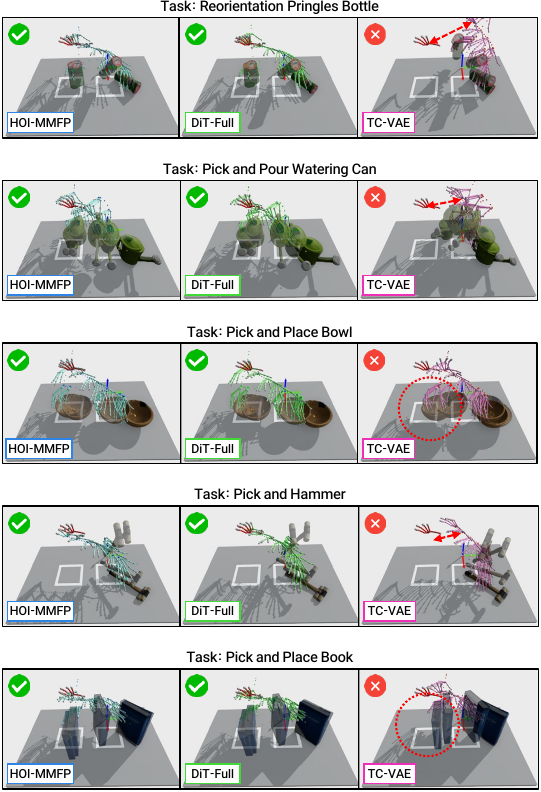}
    \caption{Generated HOI trajectories for all tasks. The red hand denotes the canonical hand pose.}
    \label{FIG:FIGURE_7}
    \vspace{-8pt}
\end{figure*}

\paragraph{Generated trajectories across all tasks.}
Figure~\ref{FIG:FIGURE_7} visualizes HOI-MMFP outputs on all five tasks.
The generated wrist--object--finger trajectories are temporally smooth,
remain faithful to the canonical initial pose enforced during
augmentation, and adapt their shape to each task's text condition.

%% figure 7 %%

\begin{figure*}[h]
    \centering
    \includegraphics[width=0.5\textwidth]{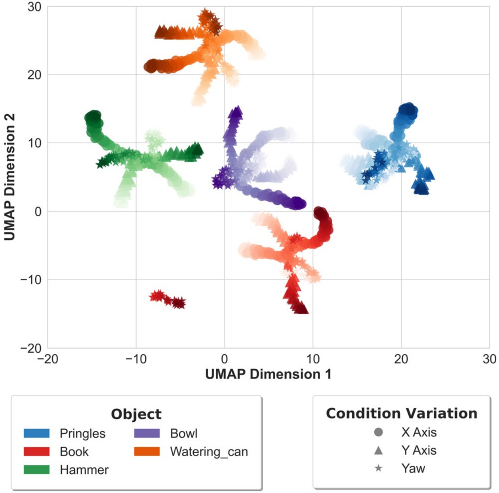}
    \caption{Conditional flow matching latent UMAP projection.}
    \label{FIG:FIGURE_8}
    \vspace{-8pt}
\end{figure*}

\paragraph{Latent organization under spatial perturbation.} 
To inspect why the latent flow generalizes across the perturbation grid of Sec.~\ref{sec:exp_motion_synthesis}, we sample $\hat{\mathbf{z}}$ under each perturbed initial object pose ($x,y$ swept at $1$\,cm resolution within $\pm 20$\,cm; yaw at $5^\circ$ within $\pm 30^\circ$) and project the latents to 2D with UMAP~\cite{mcinnes2018umap}. Figure~\ref{FIG:FIGURE_8} reveals two levels of structure. At the coarse level, the latents form distinct clusters, one per task, showing that the conditional flow first separates samples by task identity. Within each task cluster, perturbations along $x$, $y$, and yaw further trace axis-aligned variations, indicating that spatial perturbations are arranged meaningfully along the corresponding directions in latent space rather than collapsing into noise. The coexistence of global task-level clustering and locally axis-aligned spatial structure suggests that flow matching, when carried out on the learned manifold, organizes the conditional generative process in a geometrically faithful way.

%% figure 8 %%

% =====================================================================
% Appendix C
% =====================================================================
\section{Implementation Details of Residual Action RL Policy Training}
\label{appendix:APPENDIX_C}

% =====================================================================
% Appendix C.1
\subsection{Teacher Reward Design}
\label{appendix:APPENDIX_C.1}

The teacher reward is designed to preserve the kinematic structure of the synthesized HOI reference while ensuring physical feasibility under the robot's embodiment. The total reward $r_t$ is a weighted sum of five components,
\begin{equation}
r_t = \omega_{\mathrm{obj}}\, r_{\mathrm{obj}} + \omega_{\mathrm{hand}}\, r_{\mathrm{hand}} + \omega_{\mathrm{wrist}}\, r_{\mathrm{wrist}} + \omega_{\mathrm{contact}}\, r_{\mathrm{contact}} + \omega_{\mathrm{penalty}}\, r_{\mathrm{penalty}},
\end{equation}
where $d_p(\cdot, \cdot)$ and $d_r(\cdot, \cdot)$ denote the Euclidean position distance and angular rotation distance, and $a_p^{(\cdot)}, a_r^{(\cdot)}$ are positive scaling coefficients. The fingertip and wrist tracking terms, $r_{\mathrm{hand}} = \exp(-a_p^h\, d_p(\mathbf{h}_t, \hat{\mathbf{h}}_t))$ and $r_{\mathrm{wrist}} = \exp(-a_p^w\, d_p(\mathbf{w}_t, \hat{\mathbf{w}}_t)) + \exp(-a_r^w\, d_r(\mathbf{w}_t, \hat{\mathbf{w}}_t))$, are active throughout the entire episode and supervise the kinematic refinement of the synthesized reference. The object tracking reward, in contrast, is conditionally activated so that the policy is not penalized for failing to track the object pose during the reaching phase, when the hand has not yet established a viable grasp configuration. Concretely, letting $\mathbf{p}_t^{\mathrm{palm}}$ denote the palm position, $\mathbf{c}(\mathbf{o}_t)$ the object center, and $R(\mathbf{o}_t)$ the radius of the smallest object-aligned bounding sphere, we activate the term once the proximity condition $\|\mathbf{p}_t^{\mathrm{palm}} - \mathbf{c}(\mathbf{o}_t)\|_2 \leq R(\mathbf{o}_t) + \delta$ with $\delta = 10\,\mathrm{cm}$ is first satisfied at $t^{\star}$, and define
\begin{equation}
r_{\mathrm{obj}} =
\begin{cases}
\exp(-a_p^o\, d_p(\mathbf{o}_t, \hat{\mathbf{o}}_t)) + \exp(-a_r^o\, d_r(\mathbf{o}_t, \hat{\mathbf{o}}_t)), & t \geq t^{\star}, \\
0, & t < t^{\star}.
\end{cases}
\end{equation}
Once activated, the reward remains active for the rest of the episode regardless of subsequent palm-object distance, so that momentary regrasps or in-hand rolling do not deactivate object supervision. The remaining terms regulate physical interaction. The contact reward $r_{\mathrm{contact}} = \|\mathbf{b}_t\|_1$ counts the number of active binary contact sensors on the hand, encouraging stable multi-finger contact during manipulation. The penalty term $r_{\mathrm{penalty}} = p_{\mathrm{col}} + p_{\mathrm{reg}}$ aggregates a collision penalty $p_{\mathrm{col}} = -\sum_{i \in \mathcal{S}_{\mathrm{fin}}} \|\mathbf{f}_{t,i}\|_2 - \sum_{j \in \mathcal{S}_{\mathrm{bod}}} \|\mathbf{f}_{t,j}\|_2$, where $\mathcal{S}_{\mathrm{fin}}$ and $\mathcal{S}_{\mathrm{bod}}$ are the sets of contact sensors on non-fingertip hand links and arm body segments respectively, and a smoothness regularization $p_{\mathrm{reg}} = -a_v\, \|\dot{\mathbf{q}}_t\|_2$ that suppresses excessive joint velocities. The collision penalty protects the hardware and prevents the policy from exploiting spurious collisions during training, while the regularization encourages smooth, deployable motion.

% =====================================================================
% Appendix C.2
\subsection{Policy Training Details}
\label{appendix:APPENDIX_C.2}

The teacher policy, which learns residual actions for refining the synthesized HOI reference, is trained in IsaacLab~\cite{mittal2025isaac} using PPO. We use $8192$ parallel environments for teacher policy training and $4096$ parallel environments for student policy distillation. 
The reward weights and scaling coefficients used for teacher policy learning are summarized in the left panel of Table~\ref{TABLE:TABLE5}, while the PPO hyperparameters are reported in the right panel of Table~\ref{TABLE:TABLE5}. The task-space residual action integration parameters are provided in Table~\ref{TABLE:TABLE6}. Detailed observation notation and dimensions are listed in Table~\ref{TABLE:TABLE8}, and the network architecture and adaptation module configuration are summarized in Table~\ref{TABLE:TABLE7}. The domain randomization settings used during simulation training are reported in Table~\ref{TABLE:TABLE9}.

% Table 5
\begin{table}[h]
\centering
\footnotesize
\setlength{\tabcolsep}{4pt}
\renewcommand{\arraystretch}{1.08}

\begin{minipage}[t]{0.45\linewidth}
\centering
\begin{tabular}{lc}
\toprule
Symbol & Value \\
\midrule
$\omega_{\mathrm{obj}}$ & $20.0$ \\
$\omega_{\mathrm{hand}}$ & $5.0$ \\
$\omega_{\mathrm{wrist}}$ & $2.0$ \\
$\omega_{\mathrm{contact}}$ & $1.0$ \\
$\omega_{\mathrm{penalty}}$ & $0.1$ \\
$a_p^o,\; a_r^o$ & $40.0,\;0.5$ \\
$a_p^h$ & $10.0$ \\
$a_p^w,\; a_r^w$ & $20.0,\;2.0$ \\
$a_v$ & $0.3$ \\
$\delta$ & $0.10$\,m \\
\bottomrule
\end{tabular}
\end{minipage}
\hspace{2pt}
\begin{minipage}[t]{0.50\linewidth}
\centering
\begin{tabular}{lc}
\toprule
PPO Parameter & Value \\
\midrule
Value loss coef. & $1.0$ \\
Clipped value loss & True \\
Clip parameter & $0.2$ \\
Entropy coef. & $0.005$ \\
Learning epochs & $5$ \\
Mini-batches & $8$ \\
Learning rate & $5.0{\times}10^{-4}$ \\
Schedule & Adaptive \\
$\gamma$ & $0.99$ \\
$\lambda$ & $0.95$ \\
Desired KL & $0.01$ \\
Max grad. norm & $1.0$ \\
\bottomrule
\end{tabular}
\end{minipage}
\vspace{4pt}
\caption{Reward coefficients and PPO hyperparameters used for teacher policy training.}
\label{TABLE:TABLE5}
\vspace{-4pt}
\end{table}

% Table 6
\begin{table}[h]
\centering
\footnotesize
\setlength{\tabcolsep}{5pt}
\renewcommand{\arraystretch}{1.08}

\begin{tabular}{lcc}
\toprule
Parameter & Symbol & Value \\
\midrule
EMA coefficient & $\alpha$ & $0.8$ \\
Policy timestep & $\Delta t$ & $0.05$\,s \\
\midrule
Wrist position scale & $s^{w}_{p}$ & $0.02$\,m \\
Wrist rotation scale & $s^{w}_{r}$ & $5^{\circ}$ \\
Fingertip position scale & $s^{f}_{p}$ & $0.03$\,m \\
\midrule
Wrist position residual limit & $\Delta \mathbf{a}^{w,p}_{\min/\max}$ & $\pm 0.10$\,m \\
Wrist rotation residual limit & $\Delta \mathbf{a}^{w,r}_{\min/\max}$ & $\pm 40^{\circ}$ \\
Fingertip position residual limit & $\Delta \mathbf{a}^{f,p}_{\min/\max}$ & $\pm 0.05$\,m \\
\bottomrule
\end{tabular}
\vspace{4pt}
\caption{Task-space residual integration parameters.}
\label{TABLE:TABLE6}
\vspace{-4pt}
\end{table}

% Table 7
\begin{table}[h]
\centering
\footnotesize
\setlength{\tabcolsep}{5pt}
\renewcommand{\arraystretch}{1.08}

\begin{tabular}{lc}
\toprule
Parameter & Value \\
\midrule
Initial noise std. & $1.0$ \\
Intrinsic encoder hidden dims. & $[128, 64]$ \\
Intrinsic encoder output dim. & $16$ \\
Intrinsic encoder activation & ELU \\
Actor hidden dims. & $[1024, 512, 256, 128]$ \\
Critic hidden dims. & $[1024, 512, 256, 128]$ \\
Actor-critic activation & ELU \\
\midrule
Contact adaptation architecture & GRU-based temporal encoder \\
Dynamics adaptation architecture & GRU-based temporal encoder \\
GRU hidden dim. & $32$ \\
GRU layers & $2$ \\
Dynamics latent dim. & $16$ \\
Binary contact threshold & $0.5$ \\
\bottomrule
\end{tabular}

\vspace{4pt}
\caption{Network architecture.}
\label{TABLE:TABLE7}
\vspace{-4pt}
\end{table}

% Table 8
\begin{table}[h]
\centering
\footnotesize
\setlength{\tabcolsep}{4pt}
\renewcommand{\arraystretch}{1.10}

\begin{tabular}{l c p{0.50\linewidth}}
\toprule
Notation & Space & Description \\
\midrule
$\mathbf{p}^{\mathrm{arm}}_t$
& $\mathbb{R}^{23}$
& Arm proprioception, including arm joint positions $(7)$, wrist position $(3)$, wrist rotation in 6D representation $(6)$, and previous arm action $(7)$. \\

$\mathbf{p}^{\mathrm{hand}}_t$
& $\mathbb{R}^{44}$
& Hand proprioception, including hand joint positions $(16)$, fingertip positions in the world frame $(12)$, and previous hand action $(16)$. \\

$\mathbf{H}^{\mathrm{hand}}_t$
& $\mathbb{R}^{30 \times 44}$
& Hand-centric proprioceptive history over the past $30$ policy steps, used by the contact adaptation module. \\

$\mathbf{H}^{\mathrm{full}}_t$
& $\mathbb{R}^{30 \times 67}$
& Full-body proprioceptive history over the past $30$ policy steps, stacking both arm and hand proprioception. \\

$\mathbf{b}_t$
& $\{0,1\}^{11}$
& Binary contact state, including distal contacts $(4)$, medial contacts $(4)$, and proximal contacts $(3)$. \\

$\mathbf{f}_t$
& $\mathbb{R}^{33}$
& Normalized contact force, including distal contact forces $(12)$, medial contact forces $(12)$, and proximal contact forces $(9)$. \\

$\mathbf{z}^{\mathrm{dyna}}_t$
& $\mathbb{R}^{16}$
& Latent dynamics representation compressed by the dynamics encoder from object intrinsic parameters. \\

$\hat{\boldsymbol{\tau}}_t$
& $\mathbb{R}^{46}$
& Synthesized HOI reference, including target wrist pose $(3+6)$, target object pose $(3+6)$, target fingertip positions $(12)$, wrist tracking error $(3+1)$, and fingertip tracking error $(12)$. \\
\bottomrule
\end{tabular}

\vspace{4pt}
\caption{Notation and dimensions for deployable student contact and dynamics adaptation.}
\label{TABLE:TABLE8}
\vspace{-4pt}
\end{table}

% Table 9
\begin{table}[h]
\centering
\footnotesize
\setlength{\tabcolsep}{4pt}
\renewcommand{\arraystretch}{1.08}

\begin{tabular}{l l l l}
\toprule
\textbf{Parameter} & \textbf{Type} & \textbf{Distribution} & \textbf{Range} \\
\midrule
\multicolumn{4}{l}{\textbf{Robot}} \\
Mass & Scaling & Uniform & $[0.5, 1.5]$ \\
Static friction & Absolute & Uniform & $[0.2, 1.0]$ \\
Dynamic friction & Absolute & Uniform & $[0.2, 1.0]$ \\
Restitution & Absolute & Uniform & $[0.0, 0.4]$ \\
Joint stiffness & Scaling & Uniform & $[0.5, 1.5]$ \\
Joint damping & Scaling & Uniform & $[0.3, 3.0]$ \\
Joint limits & Scaling & Uniform & $[0.95, 1.05]$ \\
Joint friction & Scaling & Uniform & $[0.8, 1.2]$ \\
Armature & Scaling & Uniform & $[0.8, 1.2]$ \\

\midrule
\multicolumn{4}{l}{\textbf{Object}} \\
Mass & Absolute & Uniform & $[0.1, 1.0]$\,kg \\
Static friction & Absolute & Uniform & $[0.2, 0.6]$ \\
Dynamic friction & Absolute & Uniform & $[0.2, 0.6]$ \\
Restitution & Absolute & Uniform & $[0.0, 0.4]$ \\
COM offset $(x,y,z)$ & Additive & Uniform & $[-0.02, 0.02]$\,m \\
Uniform scale & Scaling & Uniform & $[0.95, 1.05]$ \\

\midrule
\multicolumn{4}{l}{\textbf{Desk / Partition}} \\
Static friction & Absolute & Uniform & $[0.5, 1.1]$ \\
Dynamic friction & Absolute & Uniform & $[0.5, 1.1]$ \\
Restitution & Absolute & Uniform & $[0.0, 0.4]$ \\

\midrule
\multicolumn{4}{l}{\textbf{Environment}} \\
Gravity perturbation $(z)$ & Additive & Uniform & $[0.0, 0.5]$\,m/s$^2$ \\
\bottomrule
\end{tabular}

\vspace{4pt}
\caption{Domain randomization settings used during simulation training.}
\label{TABLE:TABLE9}
\vspace{-4pt}
\end{table}

% =====================================================================
% Appendix D
% =====================================================================
\section{Details of Action Representation Baselines}
\label{appendix:APPENDIX_D}

We summarize the action representations used as baselines in Sec.~4.2. For all baselines, the synthesized HOI reference provides a reference wrist pose and hand target, denoted by $\hat{\boldsymbol{\tau}}_t$. 
The policy action is normalized to $[-1,1]$ and converted into either joint-space or task-space commands. All resulting targets are clipped to the corresponding joint or task-space limits before being sent to the inverse kinematics solver or joint controller.

\paragraph{Object-Centric Baseline Action.}
The object-centric baseline uses a mixed action representation. 
The policy predicts a wrist residual command and an absolute hand joint command:
\begin{equation}
\mathbf{u}_t =
[
\mathbf{u}^{w}_t,
\mathbf{u}^{h}_t
],
\qquad
\mathbf{u}^{w}_t \in \mathbb{R}^{6},
\quad
\mathbf{u}^{h}_t \in \mathbb{R}^{16}.
\end{equation}
The wrist residual is integrated in task space around the synthesized wrist reference:
\begin{equation}
\Delta \mathbf{a}^{w}_t
=
\mathrm{clip}
\left(
\Delta \mathbf{a}^{w}_{t-1}
+
s_w \tilde{\mathbf{u}}^{w}_t \Delta t,
\Delta \mathbf{a}^{w}_{\min},
\Delta \mathbf{a}^{w}_{\max}
\right),
\end{equation}
where $\tilde{\mathbf{u}}^{w}_t$ is the EMA-filtered wrist command. 
The refined wrist target is obtained by applying this residual to the reference wrist pose in $\hat{\boldsymbol{\tau}}_t$. 
The hand command is directly mapped to an absolute joint target:
\begin{equation}
\mathbf{q}^{h}_t
=
\mathrm{scale}
\left(
\mathbf{u}^{h}_t,
\mathbf{q}^{h}_{\min},
\mathbf{q}^{h}_{\max}
\right).
\end{equation}
The arm joint target is then obtained by solving IK from the refined wrist target.

\paragraph{Joint-Space Absolute Action.}
The joint-space absolute baseline predicts arm and hand joint commands:
\begin{equation}
\mathbf{u}_t =
[
\mathbf{u}^{a}_t,
\mathbf{u}^{h}_t
],
\qquad
\mathbf{u}^{a}_t \in \mathbb{R}^{7},
\quad
\mathbf{u}^{h}_t \in \mathbb{R}^{16}.
\end{equation}
The arm action is applied as a velocity-like update to the previous arm joint target:
\begin{equation}
\mathbf{q}^{a}_t
=
\mathrm{clip}
\left(
\mathbf{q}^{a}_{t-1}
+
s_a \mathbf{u}^{a}_t \Delta t,
\mathbf{q}^{a}_{\min},
\mathbf{q}^{a}_{\max}
\right).
\end{equation}
The hand action is mapped to an absolute hand joint target:
\begin{equation}
\mathbf{q}^{h}_t
=
\mathrm{scale}
\left(
\mathbf{u}^{h}_t,
\mathbf{q}^{h}_{\min},
\mathbf{q}^{h}_{\max}
\right).
\end{equation}

\paragraph{Joint-Space Residual Action.}
The joint-space residual baseline predicts residuals around the nominal HOI joint targets. 
The nominal targets are obtained from the synthesized HOI reference by IK:
\begin{equation}
\hat{\mathbf{q}}^{a}_t, \hat{\mathbf{q}}^{h}_t
=
\mathrm{IK}
\left(
\hat{\boldsymbol{\tau}}_t
\right).
\end{equation}
The policy predicts residual commands for the arm and hand:
\begin{equation}
\mathbf{u}_t =
[
\mathbf{u}^{a}_t,
\mathbf{u}^{h}_t
],
\qquad
\mathbf{u}^{a}_t \in \mathbb{R}^{7},
\quad
\mathbf{u}^{h}_t \in \mathbb{R}^{16}.
\end{equation}
After EMA filtering, the residuals are integrated and clipped:
\begin{equation}
\Delta \mathbf{q}_t
=
\mathrm{clip}
\left(
\Delta \mathbf{q}_{t-1}
+
s_q \tilde{\mathbf{u}}_t \Delta t,
\Delta \mathbf{q}_{\min},
\Delta \mathbf{q}_{\max}
\right).
\end{equation}
The final joint target is obtained by adding the residual to the nominal HOI target:
\begin{equation}
\mathbf{q}_t
=
\mathrm{clip}
\left(
\hat{\mathbf{q}}_t + \Delta \mathbf{q}_t,
\mathbf{q}_{\min},
\mathbf{q}_{\max}
\right).
\end{equation}
This representation constrains the policy to refine the synthesized HOI reference in joint space rather than generating the entire motion from scratch.

\paragraph{Task-Space Absolute Action.}
The task-space absolute baseline predicts task-space velocity commands for the wrist and fingertips:
\begin{equation}
\mathbf{u}_t =
[
\mathbf{u}^{w}_t,
\mathbf{u}^{x}_t
],
\qquad
\mathbf{u}^{w}_t \in \mathbb{R}^{6},
\quad
\mathbf{u}^{x}_t \in \mathbb{R}^{4 \times 3}.
\end{equation}
The wrist command is updated by integrating the predicted translational and rotational velocities:
\begin{equation}
\mathbf{a}^{w}_t
=
\mathbf{a}^{w}_{t-1}
+
s_w \mathbf{u}^{w}_t \Delta t.
\end{equation}
The fingertip targets in the wrist frame are similarly updated as
\begin{equation}
\mathbf{x}^{W}_t
=
\mathbf{x}^{W}_{t-1}
+
s_x \mathbf{u}^{x}_t \Delta t.
\end{equation}
The commanded wrist pose and fingertip positions are then converted to executable arm and hand joint commands through IK:
\begin{equation}
\mathbf{q}_t
=
\mathrm{IK}
\left(
\mathbf{a}^{w}_t,
\mathbf{x}^{W}_t
\right).
\end{equation}
Unlike the residual formulation, this baseline integrates commands from the previous task-space target rather than refining the synthesized HOI reference at each step.

% =====================================================================
% Appendix E
% =====================================================================
\section{Implementation Details of Real-World Experiments}
\label{appendix:APPENDIX_E}

\begin{figure*}[h]
    \centering
    \includegraphics[width=0.7\textwidth]{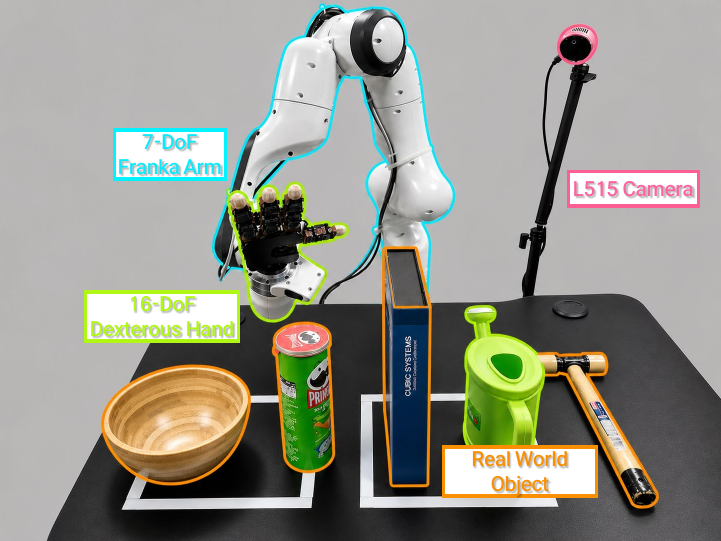}
    \caption{Real-world experimental setup.}
    \label{FIG:FIGURE_9}
    \vspace{-8pt}
\end{figure*}

\paragraph{Hardware and workspace.}
Our real-world platform consists of a Franka Panda 7-DoF arm equipped with a 16-DoF dexterous hand, and RGB-D observations are captured by an Intel RealSense L515 camera (Figure~\ref{FIG:FIGURE_9}). To evaluate the prior under realistic conditions, we use \emph{actual real-world objects} rather than 3D-printed replicas. Two white rectangles are marked on the table: the left rectangle indicates the target placement region for pick-and-place tasks, and the right rectangle defines the initial grasping region. The grasping region is sized to coincide with the in-distribution range of object placements seen during training and augmentation.

%% figure 9 %%

\paragraph{Object pose estimation.}
To obtain the object's 6-DoF pose at every step we use FoundationPose, which requires an initial object mask. At each episode start we generate this mask by combining an open-vocabulary detection from Grounding DINO~\cite{liu2024grounding} with segmentation from SAM2~\cite{ravi2024sam}; the resulting mask is supplied once to FoundationPose, which then tracks the object through the rest of the episode.

\paragraph{Policy execution.}
In simulation the policy runs at the simulator step rate of $120$\,Hz, producing task-space targets that are then quintic-interpolated. On the real robot, policy queries are issued at $20$\,Hz; their task-space targets are quintic-interpolated and streamed to \texttt{libfranka} at its $1$\,kHz real-time control loop. Because the generated reference trajectory does not always start at the robot's current configuration, we move the robot to the initial pose of the generated trajectory before each episode and roll out the policy from there.

\begin{figure*}[h]
    \centering
    \includegraphics[width=\textwidth]{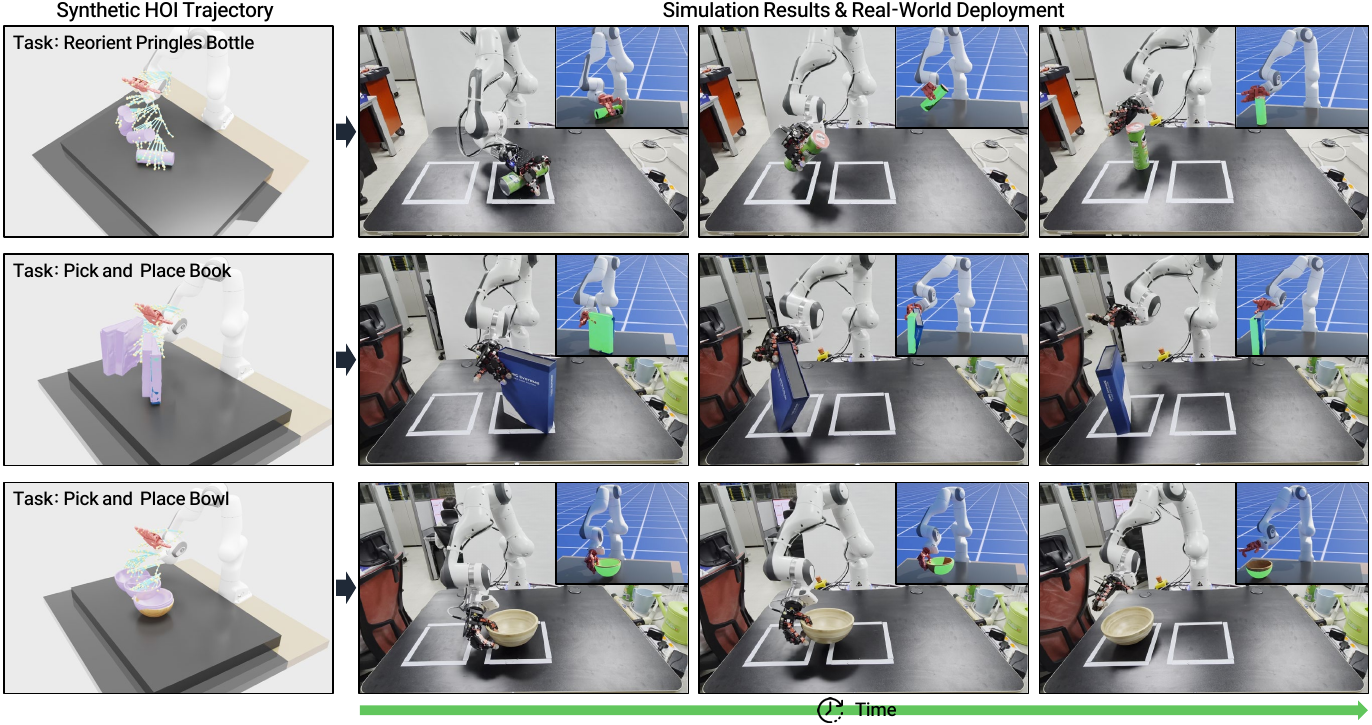}
    \caption{Real-world rollouts for the remaining tasks.}
    \label{FIG:FIGURE_10}
    \vspace{-8pt}
\end{figure*}

\paragraph{Additional qualitative results.}
Figure~\ref{FIG:FIGURE_10} shows real-world rollouts for the remaining
tasks not included in Figure~\ref{FIG:FIGURE_5}.

%% figure 10 %%

% =====================================================================
% Appendix F
% =====================================================================
\section{Joint-Limit Compliance via Residual RL}
\label{appendix:APPENDIX_F}
 
\paragraph{IK alone violates joint limits.}
We deliberately choose a Franka Panda base configuration that emulates the kinematic profile of a human arm so that wrist trajectories from the human-trained prior fall within the robot's reachable workspace. Even with this alignment, tracking the wrist reference with damped least-squares (DLS) IK frequently drives the solved joint configuration into one or more joint limits before the wrist reaches its target, causing the planner to abort. This occurs even when the wrist target itself is geometrically reachable, because DLS IK has no awareness of joint-limit headroom and will spend its redundancy arbitrarily.
 
\begin{figure*}[h]
    \centering
    \includegraphics[width=\textwidth]{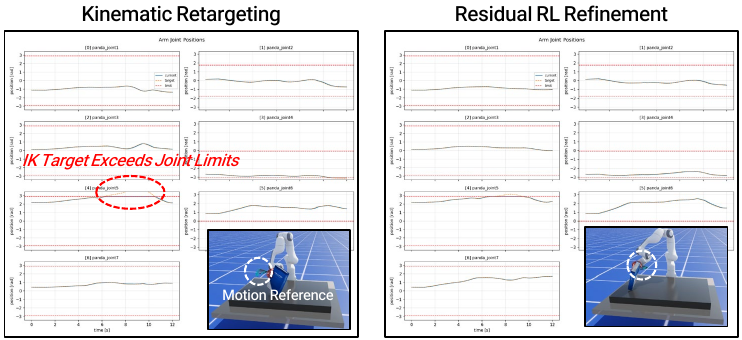}
    \caption{Joint-limit behavior under pure DLS-IK tracking
    (\emph{top}) versus residual RL (\emph{bottom}).}
    \label{FIG:FIGURE_11}
    \vspace{-8pt}
\end{figure*}

\paragraph{Residual RL restores feasibility.}
Adding the residual policy on top of the same wrist reference yields task-space targets whose IK solutions remain inside the joint-limit box throughout the rollout while still completing the task. The residual therefore acts as a learned correction that absorbs the mismatch between the human-derived reference and the robot's joint-space constraints. Figure~\ref{FIG:FIGURE_11} illustrates this comparison: the top row shows a simulation snapshot of the pure-IK rollout (left) and the corresponding per-joint trajectories over time (right), in which one or more joints saturate at their limits; the bottom row shows the residual-RL rollout, whose joint trajectories stay well clear of the limits while the task is executed successfully.
 
%% figure 11 %%

\end{document}